\documentclass[conference]{IEEEtran}
\usepackage[utf8]{inputenc}
\usepackage[nottoc]{tocbibind}

\usepackage[T1]{fontenc}
\usepackage{authblk}
\usepackage[left=0.75in, right=0.75in, top=1in, bottom=0.75in]{geometry}
\usepackage{lipsum}  
\usepackage{amsmath}
\usepackage{amssymb}

\usepackage{graphicx}
\usepackage{subcaption}

\usepackage{cite}
\usepackage{algorithm} 
\usepackage{algpseudocode} 
\usepackage{hyperref}
\usepackage{dirtytalk} 
\usepackage[linesnumbered,ruled,vlined,algo2e,resetcount]{algorithm2e}

\title{\LARGE \bf
BVR Gym: A Reinforcement Learning Environment for \\Beyond-Visual-Range Air Combat
}

\author[1,3]{Edvards Scukins }
\author[2]{Markus Klein}
\author[1]{Lars Kroon}
\author[3]{Petter Ögren}

\affil[1]{\footnotesize Aeronautical Solutions division, SAAB Aeronautics}
\affil[2]{\footnotesize Tactical Control and Data Fusion division, SAAB Aeronautics}
\affil[3]{\footnotesize Robotics, Perception and Learning Lab., Royal Institute of Technology (KTH)}

\usepackage{nomencl}
\makenomenclature

\begin{document}

\maketitle

\begin{abstract}

Creating new air combat tactics and discovering novel maneuvers can require numerous hours of expert pilots' time. Additionally, for each different combat scenario, the same strategies may not work since small changes in equipment performance may drastically change the air combat outcome. For this reason, we created a reinforcement learning environment to help investigate potential air combat tactics in the field of beyond-visual-range (BVR) air combat: the BVR Gym. This type of air combat is important since long-range missiles are often the first weapon to be used in aerial combat. Some existing environments provide high-fidelity simulations but are either not open source or are not adapted to the BVR air combat domain. Other environments are open source but use less accurate simulation models. Our work provides a high-fidelity environment based on the open-source flight dynamics simulator JSBSim and is adapted to the BVR air combat domain. This article describes the building blocks of the environment and some use cases. 

\end{abstract}


\nomenclature{\(h\)}{ Agents altitude}
\nomenclature{\(v\)}{ Agents air speed}
\nomenclature{\(v_{D}\)}{ Agents down velocity}
\nomenclature{\(\psi\)}{ Agents heading}
\nomenclature{\(\rho\)}{ Distance to the firing position }
\nomenclature{\(\nu\)}{ Initial launch velocity}
\nomenclature{\(\tau\)}{ Time since missile launch}
\nomenclature{\(\eta\)}{ Relative angle to firing position}
\nomenclature{\(\beta\)}{ Altitude of the firing position}
\nomenclature{\(MD\)}{Miss-Distance}
\nomenclature{\(a_{Head}\)}{Set heading}
\nomenclature{\(a_{Alt}\)}{Set altitude}
\nomenclature{\(a_{Thr}\)}{Set thrust}

\printnomenclature

\section{Introduction}

The nature of air combat has changed dramatically in the past half a century. Pilots can engage hostile aircraft at increasing distances due to sensors, armaments, and communication improvements. This development allows pilots to switch from within-visual range (WVR) combat to beyond-visual range (BVR)  combat. Given all these technological advancements, BVR air warfare is currently the most effective type of air combat \cite{stillion2015trends}. When pilots train, it is vital that they are exposed to a large variety of situations and opponent tactics.
Manually creating such situations and tactics can be difficult and time-consuming.
One possible way to alleviate this problem is to use Reinforcement Learning (RL).

 RL has been applied to a large variety of problem domains. A recent work \cite{kurach2020google} focuses on providing a customizable environment where the agent can learn team strategies for football. Other environments present challenges in agriculture, traffic management, and product recommendations \cite{overweg2021cropgym, zhang2019cityflow, rohde2018recogym}. In  \cite{ravaioli2022safe}, the authors noted that there is a lack of standard environments for aerospace problems. For this reason, the latter work developed an Aerospace SafeRL Framework that includes environments for aircraft formation flight and spacecraft docking in both 2D and 3D environments. 

Below, we discuss some noteworthy recent high-fidelity flight dynamics simulation engines available. AirSim \cite{shah2018airsim} is one of the more cited flight dynamics simulation environments used for AI research in the aerospace domain. It is an open-source platform built upon the game platform Unreal Engine and aims to narrow the gap between simulation and reality. Gazebo \cite{koenig2004design} is another high-fidelity simulation framework popular among robotics researchers and extends to the aerospace domain, with a focus on multi-rotor drones. X-plane \footnote{\url{https://github.com/nasa/XPlaneConnect}} from Nasa is a high-fidelity simulation environment. In \cite{pinguet2021neural}, authors used an X-plane flight dynamics engine for data collection, which was later used to train an autopilot in the form of a feed-forward neural network (FNN).  The authors of \cite{richter2022using} used the Double Deep Q-Network (DDQN) approach to train an agent for attitude control. This work used X-plane to verify the trained agents' ability to deal with complex environments. Another high-fidelity flight dynamics simulation environment is JSBSim \cite{berndt2004jsbsim}. One important recent event within the air combat domain was the DARPA AlphaDogfight Trials \footnote{\url{https://www.darpa.mil/news-events/2020-08-26}}, where teams competed against each other on algorithms that are capable of performing simulated WVR air combat maneuvering and finally competing against experienced Air Force F-16 pilots. The authors of \cite{pope2022hierarchical} participated in the trials and used RL to train an agent for this specific competition. The simulation environment used within this competition was based on the JSBSim flight dynamics engine, operating within a WVR air combat setting. 
 
Given the currently available RL environments, as seen in Table \ref{tab:sim}, there is a need for an open-source high-fidelity environment to explore tactics within the BVR air combat domain. 

\begin{table}[h!]
\centering
\caption{Overview of Simulation environments}
\label{tab:sim}
\begin{tabular}{ |c|c|c|c| } 
\hline
Simulation Environment & High Fidelity & Open source & BVR \\ 
\hline
AirSim \cite{shah2018airsim} & \checkmark & \checkmark &  \\ 
\hline
Gazebo \cite{koenig2004design} & \checkmark & \checkmark &  \\
\hline
X-plane \footnote{\url{https://github.com/nasa/XPlaneConnect}} & \checkmark & \checkmark &  \\
\hline
WUKONG \cite{piao2020beyond} & \checkmark &  & \checkmark \\
\hline
JSBSim \cite{berndt2004jsbsim} & \checkmark & \checkmark &  \\
\hline
General Motion Model\cite{yang2020evasive} &  & \checkmark &  \checkmark \\
\hline
BVRGym \cite{BVRGym} (our approach) & \checkmark & \checkmark &  \checkmark \\
\hline
\end{tabular}
\end{table}
The main contributions of this paper are as follows.
We propose a BVR air combat environment based on the high-fidelity flight dynamics simulator JSBSim. Key contributions are listed below:
\begin{itemize}
  \item it is open source
  \item it provides a set of BVR scenarios with easy integration into different RL algorithms  
  \item it provides a BVR missile model with a Proportional Navigation (PN) guidance law
  \item it gives the ability to customize and create new scenarios
\end{itemize}

The library and additional documentation are available here\footnote{\url{https://github.com/xcwoid/BVRGym}}, and Table \ref{tab:sim} above provides a comparison to related simulation environments.

\section{Background}
This framework consists of the following components: (i) Tactical units that are used to conduct BVR air combat. We use a military aircraft model and a long-range missile for this work. These models are explicitly adapted for BVR air combat. (ii) To enable the participation of units applying manually designed policies in the scenarios, we include a behavior tree implementing a simple but extendable BVR policy.  (iii) To facilitate the use of a wide range of RL algorithms, we developed a simple OpenAI-Gym-like interface \cite{brockman2016openai}.  
Additional effort has been made to make it similar to an actual BVR air combat training. In such scenarios, the teams are usually split into two groups: the blue and red teams, the red team being the adversary. In these scenarios, aircraft use radars to track the position of the opposing team since the distances implied within BVR are generally up to 100 km. To detect an aircraft at such distances, pilots use onboard radars. When the opposing team launches a missile, it is possible to detect the launch; when the missile engine ignites, it might be captured by the Infrared-Search and Track (IRST) sensors, and the detection can be associated with the tracked aircraft. In this case, estimating from which adversarial aircraft the missile has been launched is possible. Tracking the missile, on the other hand, is a much more challenging task since the missile is much smaller than the size of the aircraft that launched it. For this reason, our training environments only enable using the knowledge of from where the missile was launched and not the current position of the missile. 
Additionally, BVR air combat evolves on a slower time scale than WVR air combat, which makes typical RL rewards sparse and training for RL agents more challenging, as fewer things happen during more extended periods of time and the exploration space is large. 
As noted above, the BVR Gym enables the manual design of policies using a behavior tree (BT), a switching structure that has been shown to be optimally modular \cite{biggar2022modularity} and used to create extendable hierarchical control structures in robotics \cite{ogren2022behavior, iovino2022survey}.
 Below, we describe the basic concepts of RL and the BTs. 

\subsection{Reinforcement learning}

Reinforcement learning (RL) is a subfield of machine learning that obtains knowledge by dynamic interaction with an environment, and it offers a powerful method to train an agent for intelligent decision-making. The agent is the learning entity's representative in this scenario, and a strategy that directs the agent's decision-making process is at the core of the agent. Unlike supervised learning, where the trained model is learned on a labeled dataset, the field of RL focuses on unsupervised learning, where the model focuses on discovering patterns without explicit guidance through a set of discrete time steps. At each time step $t$, the agent perceives the environment through state representation $s_t$, can select an action $a_t$ from a set of possible actions, and receives a numerical reward $r_{t}$ provided by the environment \cite{sutton2018reinforcement}. 

Depending on the agent's condition, the obtained reward might be used to assess how good or bad a particular action was. Thus, the agent's goal is to find a $\pi(a|s)$ policy that maximizes these long-term rewards given the agent's current state. The problem can be mathematically formulated as 
$$
\pi^* = \arg\max_\pi \mathbb{E}\left(\sum_{t=0}^{\infty} \gamma^t r_{t+1} \mid \pi \right),
$$
where $\gamma$ is the discount factor to balance immediate rewards against future rewards, the expectation $\mathbb{E}$ is taken over all possible sequences of states, actions, and rewards under the policy $\pi$.

The two main methods for solving RL problems are on-policy and off-policy learning. Q-learning is one form of off-policy learning, which separates the learning policy from the policy used to investigate the provided environment \cite{sutton2018reinforcement}. On the other hand, policies are updated by on-policy reinforcement learning algorithms, like REINFORCE, based on their experiences interacting with the environment while utilizing current policy. Off-policy approaches also benefit from a broader range of experiences brought along by various policies. Finding the right balance between exploitation and exploration is crucial in reinforcement learning. To maximize immediate benefits, the agent must leverage its present knowledge while exploring to the fullest extent possible to uncover optimum behaviors. In this work, we use on-policy optimization, such as the Proximal Policy Optimization (PPO) algorithm \cite{schulman2017proximal}. One of the reasons for selecting PPO is its robustness of hyper-parameter selections for different tasks and the reported performance of the algorithm when used to address real-world issues like attitude control for fixed-wing and quad-rotor aircraft \cite{koch2019reinforcement}, \cite{bohn2019deep}.

\subsection{Behavior Trees}

Behavior Trees (BT) are a hierarchical and modular \cite{biggar2022modularity} method used in robotics, artificial intelligence, and video games to describe the policy of autonomous entities. BTs were first created by video game programmers, but they have since been used in many other fields where complex decision-making and accomplishment of tasks are crucial \cite{iovino2022survey}.

\begin{figure}[htbp]
    \centering
    \includegraphics[width=\linewidth]{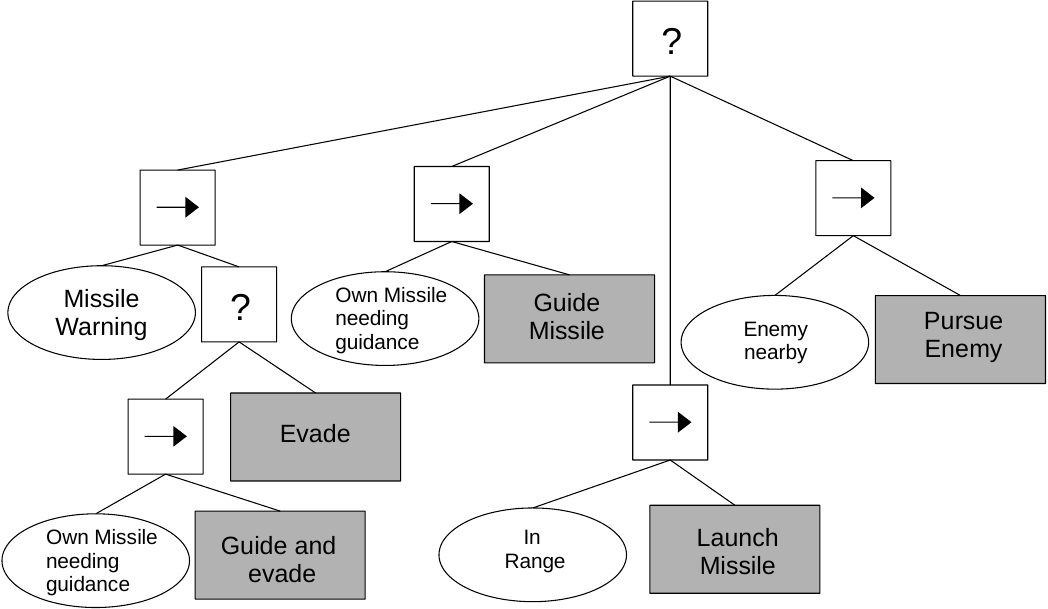}
    \caption{BT structure controlling the F16 aircraft belonging to  Team $\mathcal{R}$ .}
    \label{fig:red_bt}
\end{figure}

A behavior tree is a hierarchical structure depicting an agent's decision-making process. Unlike alternative frameworks, BTs offer a more adaptable and modular method of describing and classifying behaviors. We use BT in our work to create manual policies, and modularity is useful since BVR air combat can be decomposed into a number of complex behaviors. An example of a simple BT is illustrated in Figure \ref{fig:red_bt}. Each node in the tree structure represents a specific action or decision-making measure. The directed graph, which is made up of interconnected nodes, directs the agent through a series of decisions and actions. Control and task nodes are the two basic categories of nodes that comprise BTs \cite{colledanchise2018behavior}. Control nodes oversee the execution flow, choosing when and how to execute child nodes. In contrast, task nodes stand for discrete actions or decision points. There are two types of execution nodes (\emph{Action} and \emph{Condition}) and three primary categories of control nodes (\emph{Sequence}, \emph{Fallback}, and \emph{Parallel}); however, for this work, we do not use the \emph{Parallel}) node. Below, we will briefly describe the nodes of a given BT.

\textit{Sequences}, illustrated by a box containing the label \say{$\rightarrow$}, execute its child nodes in sequence until one fails. \textit{Fallbacks}, illustrated by a box containing the label \say{$?$}, executes its child nodes in sequence until one succeeds. \textit{Action} nodes take actions, such as avoiding a missile, engaging with an enemy, or guiding your missile to the target, and \textit{Condition} nodes check if a given condition is satisfied or not.  BTs have the advantage of being easily readable and modifiable. Designers may easily manipulate the tree structure to visualize and change the decision-making process. Because of this, BTs are a very useful tool in fields where quick iterations and prototyping are crucial.

\section{Tactical Units}
\label{sec:Background}
Two critical components of the BVR air combat are military aircraft, such as jet fighters with long-range detection systems, and long-range missiles. This section briefly describes the tactical units used within this simulation environment, namely the F-16 aircraft and the BVR missile and their properties.

\subsubsection{F-16 Aircraft}

We utilize the JSBsim F-16 flight dynamics model\footnote{\url{https://github.com/JSBSim-Team/jsbsim/tree/master/aircraft/f16}} for this training environment. While the F-16 model has its own predefined controllers to keep the inherently unstable aircraft stable, we added an additional high-level controller to adapt the unit for BVR air combat. In general, BVR air combat does not include aggressive maneuvering since pilots conserve the energy of their aircraft; for this reason, we have added an auto-pilot controller to steer the aircraft in the desired direction.  This allows the agent to set the desired heading, altitude, and throttle instead of controlling the attitude rates, reducing the RL search space and promoting faster convergence. 

Thus, if the agent chooses to set a desired direction, the lower-level controllers automatically roll the aircraft and turn it to the desired direction. Similarly, if the agent chooses to change altitude, lower-level controllers automatically stabilize the aircraft, adjust the pitch angle, and execute the maneuver to achieve the desired altitude.

\begin{figure}[htbp]
    \centering
    \includegraphics[width=\linewidth]{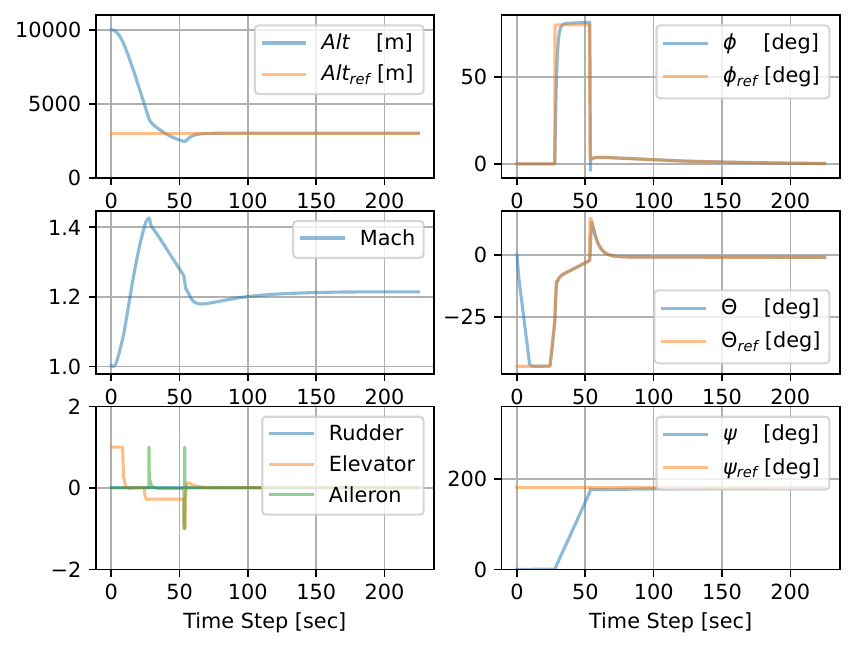}
    \caption{F-16 evasive maneuver. A drop in altitude initiates the maneuver, followed by a turn to evade the incoming missile.}
    \label{fig:f16}
\end{figure}

Since aircraft are complex systems, it is helpful to see the exact behavior of the unit of interest. For this reason, we added the possibility of studying aircraft behavior before deploying it to an RL environment. Figure \ref{fig:f16} captures aircraft dynamics while performing an evasive maneuver, including a decrease in altitude (to increase air density to promote missile deceleration) and a change in direction (to maximize the distance from the missile). 

\subsubsection{BVR Missile}

The capacity to interact with enemy fleets at great distances, up to 100 [km], is crucial in BVR air combat. For this reason, we developed a BVR Missile model. Since the exact performance of real missiles is highly classified, our missile model was inspired by performance estimates available from open sources\footnote{\url{https://en.wikipedia.org/wiki/AIM-120_AMRAAM}}. After being launched, a stage of acceleration is followed by the missile's ascent to a higher altitude. Since the air is less thick at higher altitudes, it is possible to increase the flight range by keeping a high altitude for as long as possible. To keep the missile model simple but realistic, we implemented a Proportional Navigation (PN) guidance law \cite{yanushevsky2018modern}. This law navigates the missile toward the target after reaching the desired cruise velocity and altitude. 

The PN guidance law's solution provides the desired acceleration to change the missile heading to intercept a moving target. The acceleration is converted to the appropriate velocity vector and then sent to a lower-level controller to execute the turn. Like the F-16 unit, the user can change the missile's characteristics, including its precise launching location, initial velocity, altitude, cruise altitude, and target. More tools are available to aid with aligning the missile's initial heading in the direction of the target. 

\begin{figure}[htbp]
    \centering
    \includegraphics[width=\linewidth]{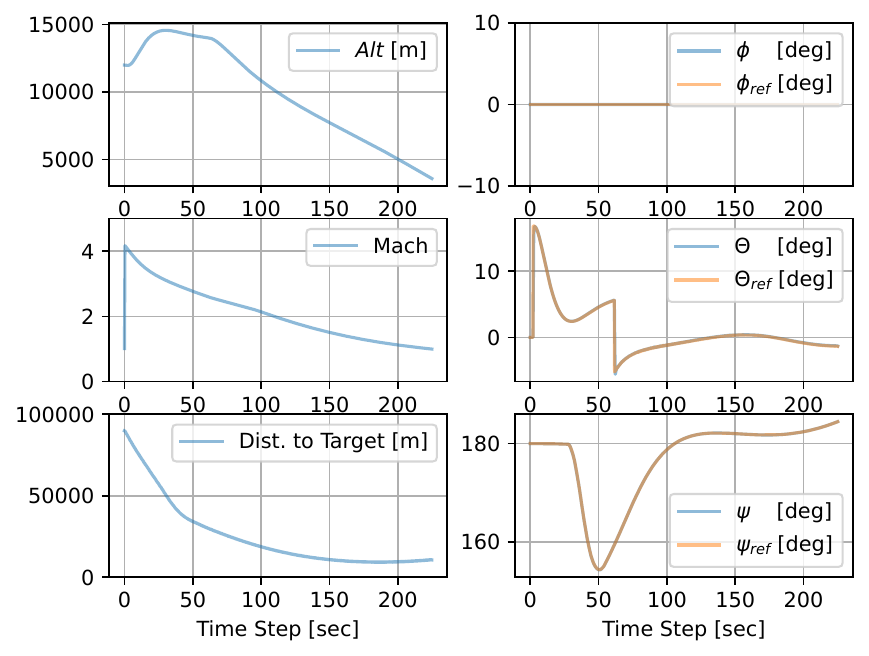}
    \caption{Missile response to F-16 evasive maneuver. Initialized by acceleration stage to Mach 4 and an ascent in altitude.}
    \label{fig:AIM}
\end{figure}

Similar to aircraft models, missiles are complex systems; hence, observing the missile's behavior before deploying it in an RL training environment is beneficial. Figure \ref{fig:AIM} depicts the missile's flight dynamics properties while the target performs an evasive maneuver away from the missile. 

\section{Scenarios}
\label{sec:Enviroments}

This section introduces a set of example BVR problems for the agent to solve. We start by introducing a problem where a single aircraft is faced with a single incoming missile. Afterward, we present a more challenging problem: the aircraft has to evade two incoming missiles launched from different locations. Finally, we present a one vs one BVR air combat scenario, where the agent needs to figure out how to defeat an adversarial aircraft.

\subsection{Evading a BVR Missile}

Consider a situation where there is one unit on each team: a single F16 aircraft on the blue team $\mathcal{B}$ and a launched missile from the red team $\mathcal{R}$. In this environment, the agent aims to find a policy that maximizes the miss distance (MD) between itself and the incoming missile.
The following observations are available to the agent at each time step
$
s_t = (h, v_{D}, v, \psi, \nu, \tau, \eta, \beta, \rho).
$ 

These observations are chosen to represent parts of a realistic air combat scenario. In such cases, when a missile is launched, the knowledge of the current missile location is usually not available since the missile is too small for the radar to detect at long ranges. However, tracking the adversary aircraft that launched it is much less complicated. Thus, an assumption can be made: if we track an aircraft and detect a sudden flash (usually representing a missile launch), we can assume that the missile was launched from the aircraft where the flash occurred. Modern military aircraft are equipped with Missile Approach Warning systems (MAW) that may detect the flash associated with the missile launch. 

When a missile launch has occurred, pilots tend to perform an evasive maneuver to evade the incoming missile. In BVR air combat, super maneuverability is not a must; thus, in most cases, complex maneuvers are not used in order to preserve aircraft momentum. For this reason, the action space can be broken down into the following actions.  
$
a_t = (a_{Head}, a_{Alt}, a_{Thr}).
$
At each time step, the agent receives a reward $r_t = R(s_t,a_t)$ that is $r_t = 0$ except for the last step, when the missile has either hit the target or has depleted all fuel and speed so that it cannot get closer to it. In the case of a successful missile evasion, the agent then receives a reward $r_T > 0$ equivalent to the MD, which is equal to the smallest encountered distance between the agent and the missile.
If the missile hits the agent, this distance is zero, resulting in a zero reward. 

\begin{figure}[htbp]
    \centering
    \includegraphics[width=\linewidth]{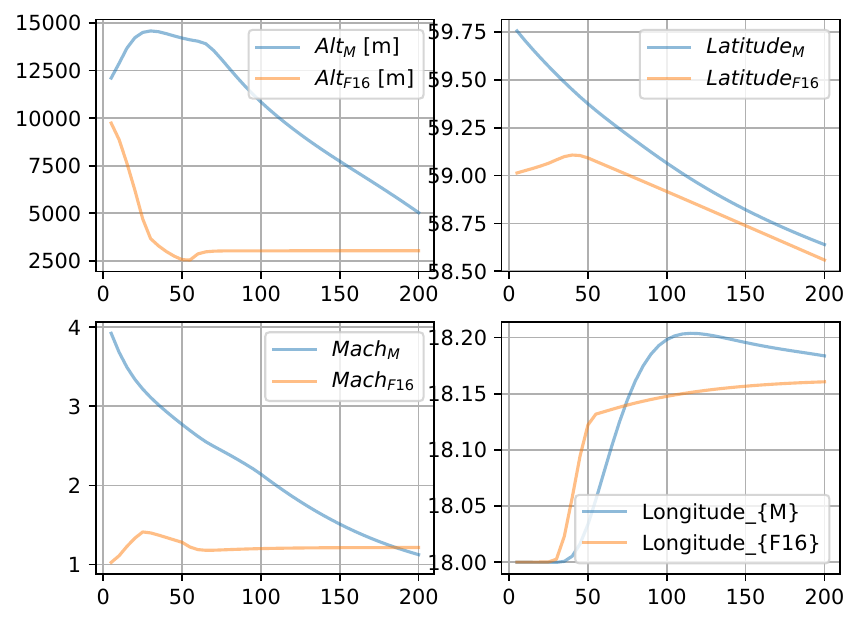}
    \caption{Missile response to F-16 evasive maneuver. Initialized by acceleration stage to Mach 4 and an ascent in altitude.}
    \label{fig:env}
\end{figure}

\subsection{Evading two BVR Missiles}

We now consider an agent subjected to two incoming missiles launched simultaneously from different locations. This is an interesting problem to study since pilots in such situations must carefully consider their options when making maneuvers to avoid missiles. Focusing purely on evading one of the two missiles can expose the aircraft to the other one. Thus, multiple threats require splitting the focus and finding solutions that can avoid both. Additionally, more missiles usually mean more maneuvering, which makes controlling energy depletion challenging. 

The observation assumptions about the two missile cases are similar to the single missile case. The agent has access to both launch locations for missiles M1 and M2 and the state observation
$
s_t = (h, v_{D}, v, \psi, \nu^{M1}, \tau^{M1}, \eta^{M1}, \beta^{M1}, \rho^{M1}, \nu^{M2}, \tau^{M2}, \eta^{M2}, \\ \beta^{M2}, \rho^{M2}) 
$, with the same action space available as in the problem with a single incoming missile. The reward is provided in a similar manner, where it is proportional to the smallest encountered distance between the agent and any of the two missiles. At every time step, the agent receives a reward $r_t^{M1} = R(s_t,a_t)$ and $r_t^{M2} = R(s_t,a_t)$ which both are equal to $0$ except for the last step. In the case of a successful missile evasion, the agent then receives a reward $min(r_{T}^{M1}, r_{T}^{M2} ) > 0$ equivalent to the MD, which is equal to the smallest encountered distance between the agent and the incoming missiles.


\subsection{BVR DogFight}
\label{sec:env_dog}
The aircraft's radar and other sensors play a critical role in the efficacy of BVR engagements. Challenges may arise from a sensor's limited precision, range, or sensitivity to electronic countermeasures. To simplify this, we consider that the location of the enemy aircraft is known without any interference. One of the crucial decisions that have to be made during air combat is the timing of when to fire and when to hold fire. Engaging in combat too soon could disclose the aircraft's location, while waiting too long could make you the target of the initial attack. BVR engagements also involve controlling the aircraft's altitude and speed to maximize missile performance after launch. 


This scenario aims to find possible counterattack strategies against an enemy with known behavior. 
To capture the enemy policy, we use BTs, which have shown to be an effective tool for creating sophisticated behaviors and have been used to define behaviors within the air combat domain \cite{iovino2022survey}. Since BVR combat is rarely observable in practice, with low availability of historical data, much of its possibilities must be assessed through simulation \cite{dantas2021engagement}. For this reason, you might want to study potential solutions to a given adversary behavior.

We have equipped the adversary aircraft with a BT that dictates the actions to take during air combat. The strategy is visualized in Figure \ref{fig:red_bt}. The main focus of this BT is to prioritize one's own safety; for this reason, missile evasion tactics are placed on the left-hand side of the tree, while offensive tactics are located on the right-hand side. The following state observation is available to the agent $s_t = \\ (
\rho^{\mathcal{B}\mathcal{R}},
\nu^{\mathcal{B}\mathcal{R}},
v^{\mathcal{B}},
h^{\mathcal{B}},
\psi^{\mathcal{B}},
v^{\mathcal{R}},
h^{\mathcal{R}},
\rho^{\mathcal{B}\mathcal{M}}_{0},
\nu^{\mathcal{B}\mathcal{M}}_{0},
v^{\mathcal{M}}_{0},
h^{\mathcal{M}}_{0})$.


The components $(\rho^{\mathcal{B}\mathcal{M}}_{0}, \nu^{\mathcal{B}\mathcal{M}}_{0}, v^{\mathcal{M}}_{0}, h^{\mathcal{M}}_{0})$ indicating observations with respect to the launch location. If there are no active missiles launched by the $\mathcal{R}$ team aircraft, then $(\rho^{\mathcal{B}\mathcal{M}}_{0}, \nu^{\mathcal{B}\mathcal{M}}_{0}, v^{\mathcal{M}}_{0}, h^{\mathcal{M}}_{0})$ are equivalent to the $(\rho^{\mathcal{B}\mathcal{R}}, \nu^{\mathcal{B}\mathcal{R}}, v^{\mathcal{R}}, h^{\mathcal{R}})$, indicating that the missile is located at the same place as the aircraft carrying it. The following action space is available to the agent $a_t = (a_{Head}, a_{Alt}, a_{Thr}, a_{l})$, with an additional change being the missile launch capability $a_{l}$.

In this environment, the agent receives a reward $r_t = R(s_t,a_t)$, which is equal to $0$ except for the last step. In the case of a successful adversary kill, the agent then receives a reward $r_{T} = 1$. If the adversary successfully shoots down the agent or the agent hits the ground, or the scenario time runs out, then a reward of $r_{T} = -1$; is provided.


\section{Numerical results}
\label{sec:NumericalResults}
This section presents results obtained from training an agent in different environments. We first consider the environment where the agent is faced with one and two incoming missiles, followed by a BVR air combat scenario.  

\subsection{One and two missile scenario}
\label{sec:missile}
\begin{figure}[htbp]
    \centering
    \includegraphics[width=\linewidth]{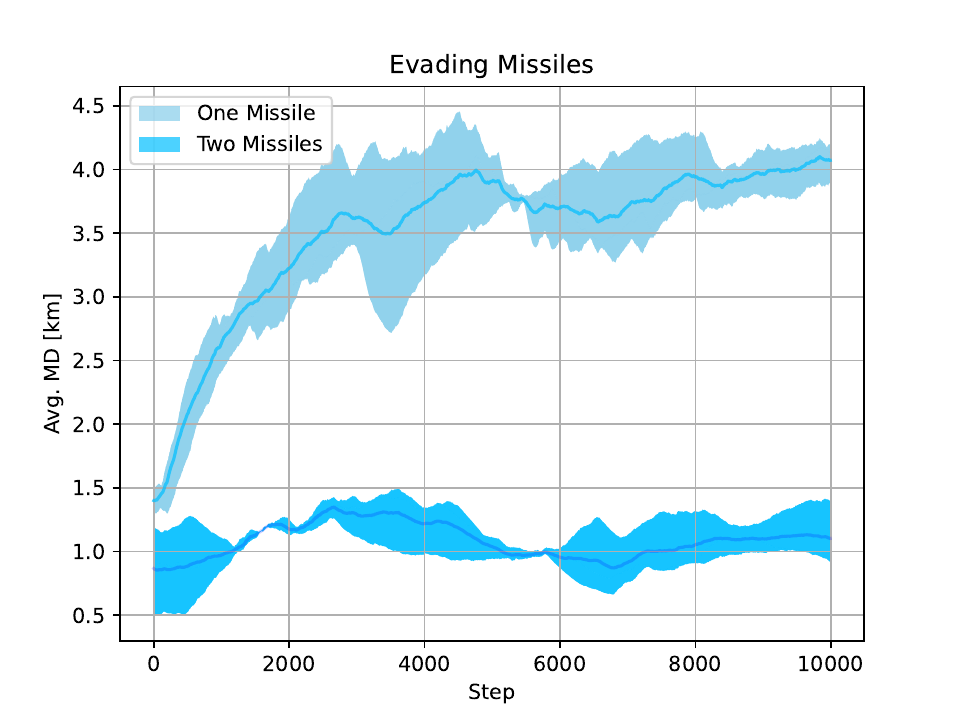}
    \caption{ Training results for evading one and two missile scenarios. The Y-axis shows the average distance by which the missile misses its target in kilometers.}
    \label{fig:one_two_RL}
\end{figure}

Figure \ref{fig:one_two_RL} shows the values obtained from the training with both one and two incoming missiles. The initial conditions of the agent and the missiles are both randomized. Both the missile's initial launch conditions and the agent's initial state are presented in Table \ref{tab:init_val}.
\begin{table}[hbt!]
\centering
\caption{Table of initial conditions.}
\begin{tabular}{p{3.5cm} l   r }
\hline
Parameter & Value \\ 
\hline
Initial Velocity: Agent  & $300 - 365 $ [m/s]  \\
Initial Velocity: M1,M2   & $280 - 320 $ [m/s]  \\ 
Initial Altitude: Agent  & $6000 - 10000 $ [m]  \\
Initial Altitude: M1,M2   & $9000 - 11000 $ [m]  \\ 
Firing Distance: M1,M2   & $40 - 80$ [km] \\ 
Initial heading: Agent  & $0-360 $ [deg] \\
Initial pitch/roll: Agent  & $0 , 0$ [deg] \\
\hline
\end{tabular}
\label{tab:init_val}
\end{table}

A closer look at Figure \ref{fig:one_two_RL} reveals that the agent typically achieves a greater separation with a single missile scenario because it is easier to establish a strategy without conflicting objectives than in a two-missile scenario. The trained model was able to improve the evasive maneuver in both situations, increasing the distance between the missile and the aircraft and preventing a missile hit. The primary tactic employed is the same as what pilots practice in training: lowering the aircraft's altitude to be surrounded by denser air, which promotes the missile to slow down and travel in the opposite direction from its launch point. Air combat problems with high-fidelity models typically require large computational budgets, as in \cite{pope2022hierarchical}. We decrease the search space by setting the throttle to maximum, leaving the agent with the action space $a_t = (a_{Head}, a_{Alt})$. Such changes, in combination with a lower-level flight controller, reduce the need for such a budget. Additional changes have been made compared to our previous work in \cite{scukins2023enhancing} to speed up convergence, and all the parameters can be found in \cite{BVRGym}. One day's worth of computing is needed to solve the problem using an Intel Core i7-8700CPU@3.20GHz, ten CPUs operating in parallel, and one NVIDIA GeForce GTX 1080 GPU for neural network optimization. Changing the default settings, such as step time, the ability to control thrust, or increasing observation space, may significantly increase computation time. 

\subsection{BVR Dogfight}

\begin{figure}[htbp]
    \centering
    \includegraphics[width=\linewidth]{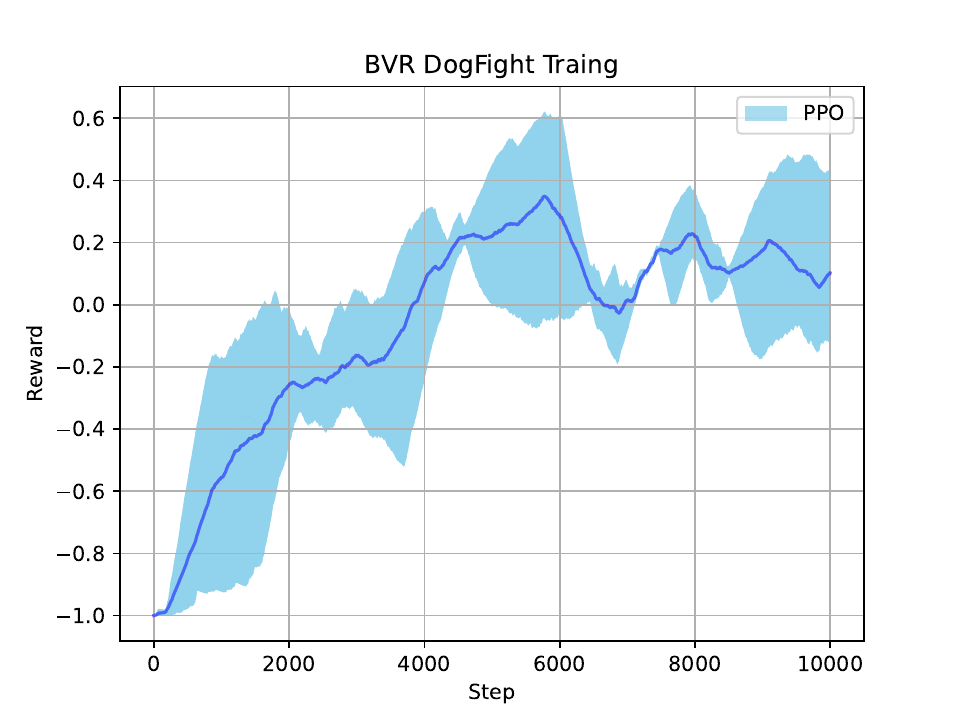}
    \caption{ One vs One BVR air combat training results. The Figure illustrates the average obtained reward (Accumulated Reward) of ten in parallel executed episodes.}
    \label{fig:Dog_RL}
\end{figure}

In this scenario, we let two aircraft face each other with two BVR missiles each. Each scenario lasts up to 16 minutes, slightly longer than in a related work done by \cite{dantas2021engagement} where the authors used a 12-minute time cap. The agent's goal in this scenario is to explore tactical policies to defeat the opposing aircraft that behaves according to the BT formulation in Figure~\ref{fig:red_bt}. The training progress results are shown in Figure~\ref{fig:Dog_RL}. 

\begin{figure}[htbp]
    \centering
    \includegraphics[width=\linewidth]{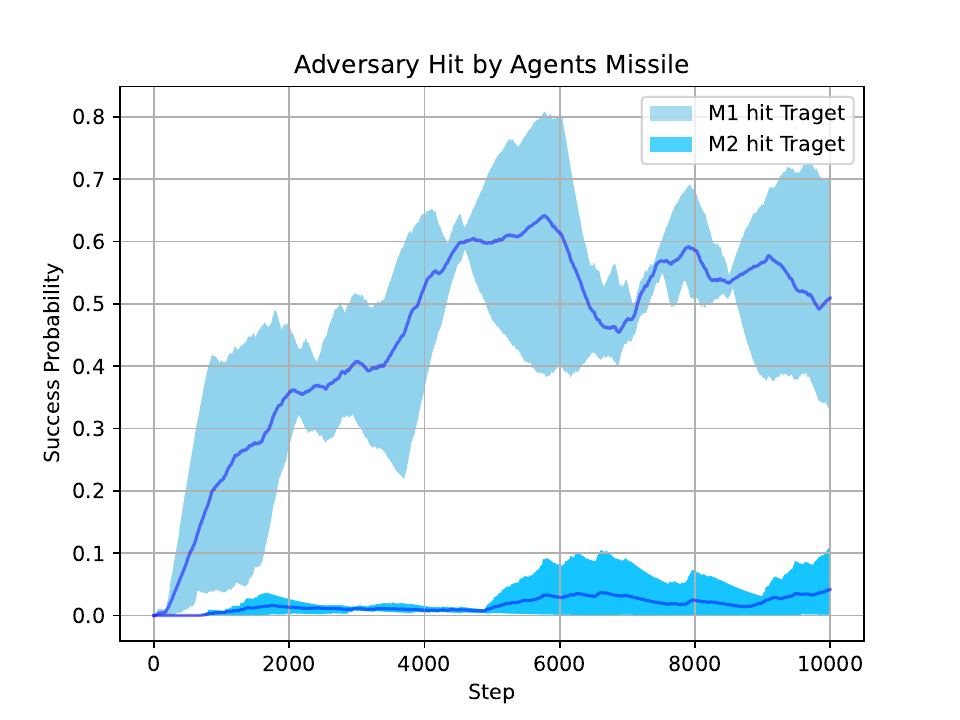}
    \caption{An illustration of the success rate of utilizing the first missile.}
    \label{fig:aim}
\end{figure}

\begin{figure}[htbp]
    \centering
    \includegraphics[width=\linewidth]{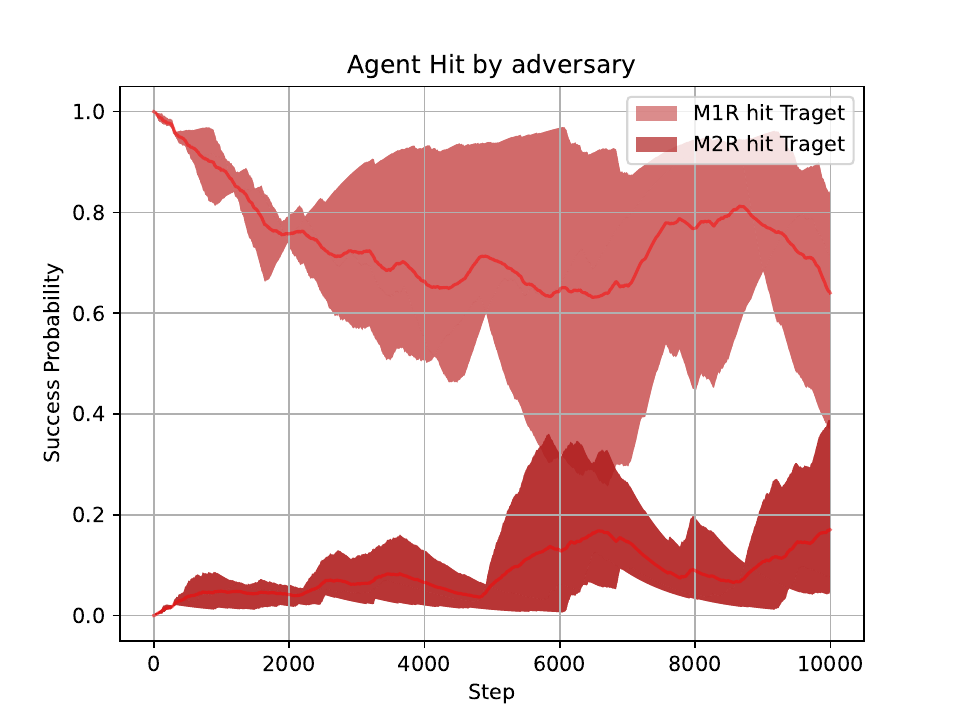}
    \caption{An illustration of the success rate of utilizing the second missile.}
    \label{fig:aimr}
\end{figure}

Upon examining Figure \ref{fig:Dog_RL}, we can observe that the agent is consistently shot down by the adversary at the beginning. Following a training period, we observe that the agent begins to behave better and wins more battles than it loses. A closer examination of the issue may also be done by examining how weapons are used in Figures \ref{fig:aim} - \ref{fig:aimr}. Figure \ref{fig:aim} shows the frequency of the agent's missile usage. The missiles were not fired since the agent did not first approach the enemy to get within firing range. Following a period of training, we observed that the first missile was launched periodically and then, in the later phases of the training, a second missile. Looking at the adversary's actions, we can observe that it consistently, and with a high likelihood of success, used the first missile successfully. But when the agent became adept at dodging the first missile, the enemy began to use the second missile even more frequently.
 
BVR air combat, incorporating several tactical units, often leads to computationally intensive problems. To speed up the process, we introduced the following steps: (i) reduce the search space concerning the previous environments, as described in Section \ref{sec:env_dog}, (ii) let the agent make decisions once in 10 seconds, (iii) no variation in starting position for the agent and the adversary (iv) the missile launch is automated. This implies that the missile will be launched automatically when the launch conditions are satisfied. The main reason is to limit exploration space since missile launch action is a one-time action lasting for approximately 2-4 minutes of simulation time, depending on the distance. If the missile launch is made outside its range, the missile can be considered lost. The same computational resources were used in this scenario as in  Section \ref{sec:missile} above, and all the parameters and the code can be found in \cite{BVRGym}.

\section{Conclusion}
\label{sec:Conclusion}
In this work, we presented a high-fidelity environment to investigate tactics with Beyond Visual Range air combat. We showed three case study scenarios to explore different aspects of BVR air combat. We also suggested some configuration parameters that enable users with limited computational resources to investigate the problems.

\section*{Acknowledgment}
The authors gratefully acknowledge funding from Vinnova, NFFP7, dnr 2017-04875.

\bibliographystyle{unsrt}
\bibliography{bibliography}

\end{document}